\documentclass{article} %
\usepackage{iclr2027_conference,times}

\usepackage{amsmath,amsfonts,bm}

\def\eqref#1{equation~\ref{#1}}

\def\1{\bm{1}}

\DeclareMathAlphabet{\mathsfit}{\encodingdefault}{\sfdefault}{m}{sl}
\SetMathAlphabet{\mathsfit}{bold}{\encodingdefault}{\sfdefault}{bx}{n}

\usepackage{url}
\usepackage{graphicx}
\usepackage{booktabs}
\usepackage{makecell}
\usepackage{placeins}
\usepackage{xcolor,colortbl}
\colorlet{methodrowgray}{black!8}
\usepackage[T1]{fontenc}
\usepackage{float}
\usepackage{amssymb}
\usepackage{wrapfig}
\usepackage[hidelinks]{hyperref}

\newcommand{\papertablestyle}{%
  \small
  \setlength{\tabcolsep}{4pt}%
  \renewcommand{\arraystretch}{1.05}%
  \setlength{\heavyrulewidth}{0.8pt}%
  \setlength{\lightrulewidth}{0.35pt}%
  \setlength{\aboverulesep}{1.5pt}%
  \setlength{\belowrulesep}{2pt}%
}

\newcommand{\centeredoverallscore}[2]{%
  \makebox[1.8em][r]{#1}%
  \makebox[1.7em][l]{%
    \if\relax\detokenize{#2}\relax\else%
      $^{\uparrow #2}$%
    \fi}%
}

\title{StateSkill-Agent: An Effective and Efficient Embodied Agent via State-conditioned Skill}

\author{%
\textbf{Changxu Wei$^{1,6}$, Jingyi Zhang$^{2}$, Zhiyu Zheng$^{1}$, Tianyi Lin$^{1}$, Yixian Zhang$^{3}$,}\\
\textbf{Yunpeng Qing$^{4,6}$, Jingyuan Liu$^{5,6}$, Zhi Hou$^{6}$, Jiaxing Huang$^{1}$\thanks{Corresponding author.}}\\[3pt]
{\normalfont $^{1}$ The Hong Kong Polytechnic University, $^{2}$ Nanyang Technological University,}\\
{\normalfont $^{3}$ Tsinghua University, $^{4}$ Zhejiang University, $^{5}$ Fudan University, $^{6}$ ACE Robotics}
}

\hypersetup{
  pdftitle={StateSkill-Agent: An Effective and Efficient Embodied Agent via State-conditioned Skill},
  pdfauthor={Changxu Wei, Jingyi Zhang, Zhiyu Zheng, Tianyi Lin, Yixian Zhang, Yunpeng Qing, Jingyuan Liu, Zhi Hou, Jiaxing Huang}
}
\iclrfinalcopy
\begin{document}

\setcounter{footnote}{1}
\maketitle
\fancyhead{}
\lhead{Preprint}

\begin{abstract}
In this work, we aim to develop an embodied MLLM agent that understands and solves embodied tasks by learning, abstracting, and reusing transferrable and generalizable behavioral patterns.
To this end, we propose StateSkill, a new skill representation for embodied agents, which introduces the concept of ``finite-state machines'' into skill modeling for effective and efficient skill construction and learning.
The core idea of StateSkill is to model embodied interaction experiences as a finite-state machine, where a skill is organized around a set of task-progress states, state transitions, and state-conditioned guidance, enabling embodied agents to dynamically adjust their skills as the task progresses while keeping skill contexts compact and stage-relevant.
We further develop StateSkill-Evolve, an experience-driven evolution mechanism that iteratively refines StateSkill through retrospective analysis of new successful and failed state-guided trajectories.
Upon this, we introduce StateGRPO, which integrates StateSkill evolution and model optimization via state-conditioned GRPO into a unified iterative learning loop, enabling state-conditioned skills and model capabilities to co-evolve and reinforce each other.
Extensive experiments demonstrate the effectiveness of our proposed methods and trained model (i.e., StateSkill-Agent) across various embodied tasks.

\end{abstract}

\section{Introduction}
\begin{figure}[t]
    \centering
    \includegraphics[width=\linewidth]{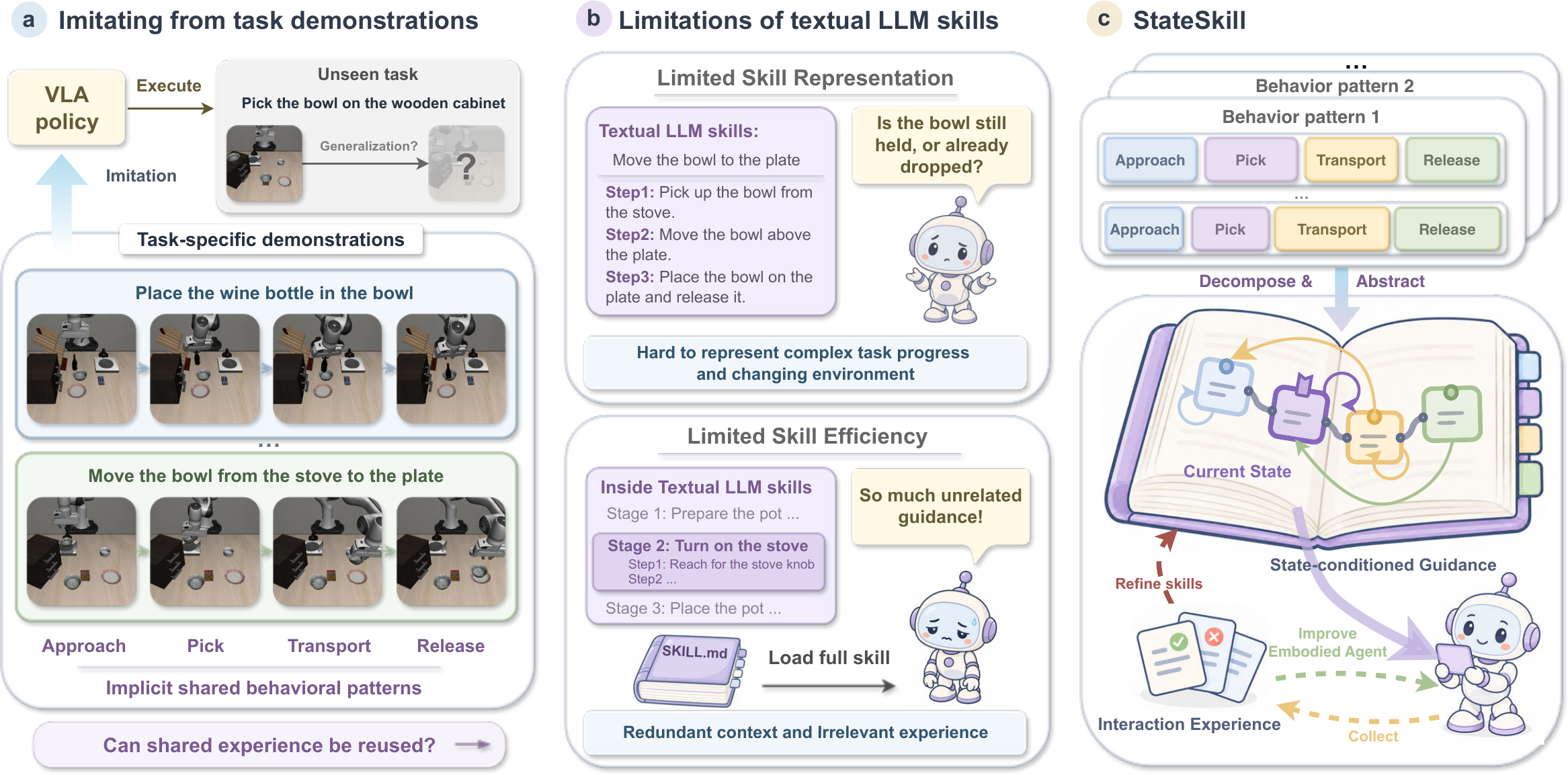}
    \caption{
    \textbf{Motivation.}
    \textbf{(a)} Imitating from task-specific demonstrations offers limited transferability and generalization.
    \textbf{(b)} Textual LLM skills face two key challenges for embodied agents: \textbf{\textit{Limited Skill Representation}} in representing complex behavioral experiences in interactive physical environments; and \textbf{\textit{Limited Skill Efficiency}} due to loading entire skill descriptions that introduce redundant and irrelevant context.
    \textbf{(c)} StateSkill organizes shared behavioral patterns into finite-state machines as skills and dynamically provides state-conditioned guidance as tasks progress, while new interaction experiences enable StateSkill and the model to co-evolve and mutually reinforce each other.
    }
    \label{fig:stateskill-motivation}
\end{figure}

Embodied models like Vision-Language-Action (VLA) models have recently emerged as a promising paradigm for building general-purpose agents for embodied AI by mapping visual observations and language instructions directly to robot actions~\citep{zitkovich2023rt2,kim2024openvla}. 
Despite rapid progress driven by large-scale collections of robot demonstrations~\citep{openx2024,ghosh2024octo}, current VLA methods remain highly dependent on the coverage and diversity of available trajectories, and their performance can degrade substantially when demonstrations are limited~\citep{lin2025data}. 
A key reason lies in that these methods predominantly learn by directly imitating task-specific demonstrations, with few explicit mechanisms for identifying, abstracting, and reusing behavioral regularities shared across demonstrations of different tasks as in Figure~\ref{fig:stateskill-motivation} (a). 
Consequently, the resulting models may overfit to training tasks, leading to limited transferability and generalization, particularly in data-scarce scenarios.

In this work, we aim to develop an embodied MLLM agent that understands and solves embodied tasks by learning, abstracting, and reusing transferrable and generalizable behavioral patterns.
Recent advances in NLP, such as Anthropic's Claude~\citep{anthropic2025skills}, have shown strong potential for enabling large language models (LLMs) to abstract past experiences into compact behavioral knowledge and utilize such knowledge to solve new tasks, substantially improving experience transfer and generalization~\citep{wang2024voyager,zhao2024expel,xia2026skillrl}. A core design underlying these advances lies in the concept of ``skill'':  it summarizes generalizable behavioral patterns from successful or failed trajectories, capturing what actions are effective, under what conditions they should be applied, and what mistakes should be avoided, which together can subsequently guide the LLM in solving relevant but unseen tasks \citep{xia2026skillrl,ma2026skillgen}. 

An intuitive idea is to directly apply these LLM skill methods to summarize and reuse behavioral knowledge for embodied MLLM agents, which, however, does not perform well. As illustrated in Figure~\ref{fig:stateskill-motivation} (b), we attribute this limitation primarily to two key observed challenges. 
(1) Skill Representation: Most existing LLM skills are typically represented in textual form while embodied agents are required to perceive, understand and make long-horizon decisions in complex and interactive physical environments, where such text-based skills are often insufficient to represent the complex behavioral experiences and task states arising from agent-environment interactions (e.g., spatial relationships, agent states, and environment states), thereby leading to limited performance.
(2) Skill Efficiency: Most existing LLM skill paradigms commonly load an entire skill description at once. For embodied tasks involving multiple execution stages and repeated physical interactions, this can introduce excessively long contexts containing substantial redundant or stage-irrelevant information, which increase computational overhead and may introduce distracting or conflicting guidance, particularly for embodied multimodal agents with limited long-context understanding capabilities.

To tackle these challenges, we propose StateSkill, a new skill representation for embodied agents, which introduces the concept of ``finite-state machines'' into skill modeling for effective and efficient skill construction and learning as illustrated in Figure~\ref{fig:stateskill-motivation} (c).
The core idea of StateSkill is to model embodied interaction experiences as a finite-state machine, where a skill is organized around a set of task-progress states, state transitions, and state-conditioned guidance.
Specifically, StateSkill first decomposes embodied trajectories of similar tasks into multiple progress stages, from which it abstracts a set of progress states and their associated state transitions, and further summarizes state-conditioned skills. Each state is equipped with (a) state-conditioned language guidance describing how to accomplish the current stage or subtasks, (b) a state transition gate specifying when and how the agent should transition to next states, and (c) state-conditioned action priors modeled from embodied trajectories to guide action prediction. 
By explicitly conditioning skills on progress states, StateSkill enables an embodied agent to track the continuously changing states of the agent, environment, and task, determine state transitions based on the current state and newly observed information, and dynamically adjust its skills in a state-conditioned manner as the task progresses.

In this way, StateSkill enables effective and efficient skill construction for embodied agents. 
(1) The finite-state representation enables StateSkill to organize embodied interaction experiences into explicit task-progress states and transitions, equipping each state with state-conditioned language guidance, a state transition gate and trajectory-derived action priors. 
Thus, it enables skills to represent complex behavioral experiences and task states arising from agent-environment interactions in a structured, state-conditioned manner, allowing embodied agents to perceive, understand and make long-horizon decisions effectively in complex and interactive physical environments.
(2) The state-conditioned guidance and state transition gates enable StateSkill to load the language guidance associated with the current progress state and dynamically adjust it as the task progresses, with transitions determined by the current state and newly observed information. This effectively keeps skill contexts compact and stage-relevant, reducing redundant information and the risk of distracting or conflicting guidance and thus lowering computational overhead and easing the long-context processing burden on embodied multimodal agents.

Furthermore, we extend StateSkill with an iterative experience-driven evolution mechanism, termed StateSkill-Evolve, enabling it to continuously evolve by learning from new state-guided experiences beyond its initial construction from pre-generated trajectories. StateSkill-Evolve retrospectively analyzes new successful and failed trajectories, identifies strengths and diagnoses weaknesses in the current StateSkill, and revises its state-conditioned guidance, state transition gates, and action priors.

In addition, we introduce StateGRPO, which integrates StateSkill evolution and model optimization into a unified iterative learning loop. 
At each iteration, the current StateSkill guides the model to generate embodied trajectories with state-conditioned guidance, transition control, and action priors, while the generated trajectories are then used to optimize the model via state-conditioned GRPO. %
The updated model then generates new embodied trajectories with enhanced capabilities, which in turn provide fresh experience and feedback for further StateSkill evolution.
By alternating between skill evolution and model optimization, StateSkill and the model progressively co-evolve, enabling structured behavioral knowledge and learned model capabilities to mutually reinforce one another. 
With StateSkill, StateSkill-Evolve and StateGRPO, we train our model, StateSkill-Agent.

The main contributions of this work are fourfold.
First, we introduce the concept of finite-state machines into skill modeling and learning for embodied agents and propose StateSkill, which structures reusable behavioral knowledge into state-conditioned skills for effective and efficient embodied execution.
To the best of our knowledge, this is the first work that explores finite-state machines for skill modeling and learning in embodied MLLM agent.
Second, we develop StateSkill-Evolve, an experience-driven evolution mechanism that iteratively refines StateSkill through retrospective analysis of new state-guided trajectories.
Third, we introduce StateGRPO, which unifies StateSkill evolution and model optimization via state-conditioned GRPO in an iterative loop, enabling skills and model capabilities to co-evolve.
Fourth, extensive experiments demonstrate the superiority of our proposed methods and trained model (i.e., StateSkill-Agent) on various benchmarks.

\begin{figure}[t]
    \centering
    \includegraphics[width=\linewidth]{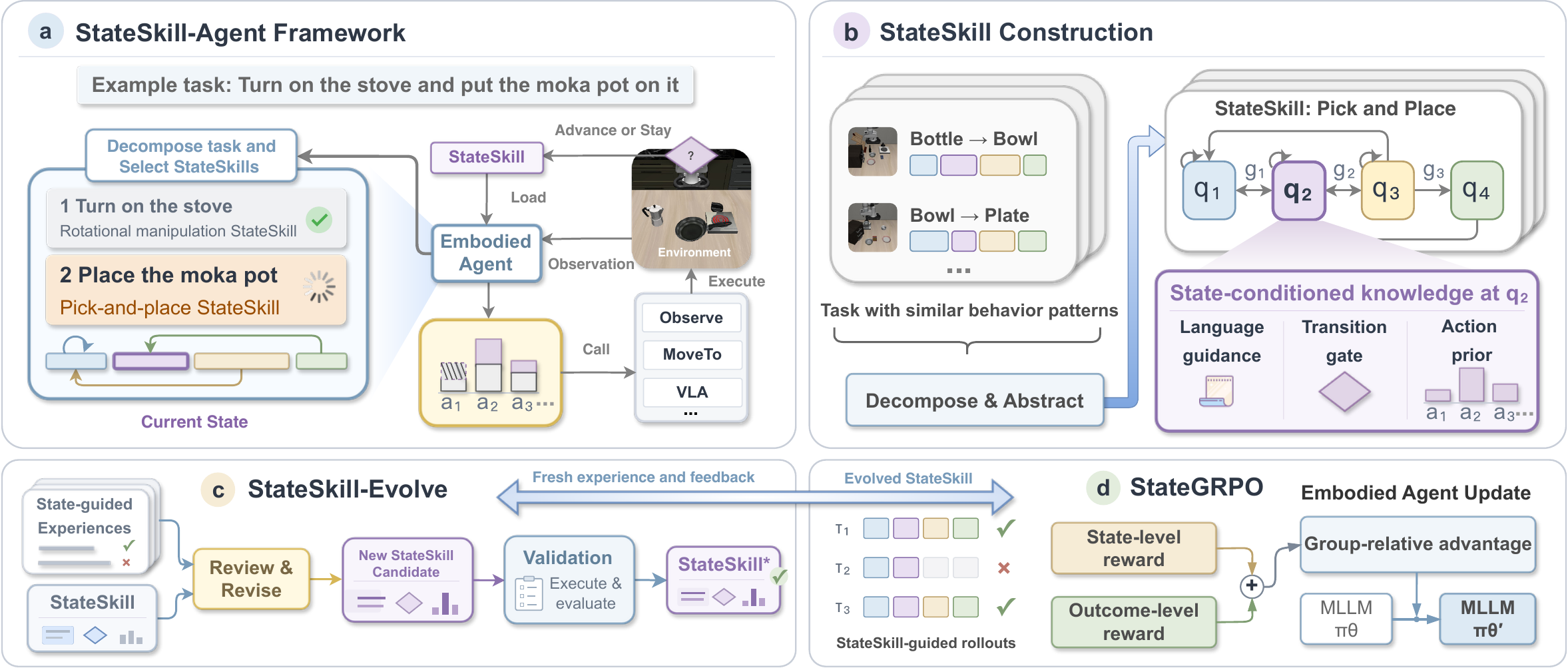}
    \caption{
    \textbf{Overview.}
    \textbf{(a)} StateSkill guides embodied execution through explicit state transitions and progressively loaded state-conditioned guidance, adapting to task progress and environment states.
    \textbf{(b)} StateSkill abstracts shared behavioral patterns from embodied trajectories into a finite-state machine, equipping each task-progress state with language guidance, a transition gate, and trajectory-derived action priors.
    \textbf{(c)} StateSkill-Evolve iteratively refines StateSkill by reviewing new successful and failed state-guided trajectories and validating candidate revisions.
    \textbf{(d)} StateGRPO integrates StateSkill evolution and state-conditioned embodied agent optimization into a unified iterative loop, enabling StateSkill and model capabilities to co-evolve.
}
    \label{fig:stateskill-framework}
\end{figure}

\section{Related Work}

\subsection{Embodied Model}
Embodied models aim to enable agents to perceive environments and execute physical actions based on observations and instructions~\citep{brohan2022rt,chi2025diffusion,driess2023palme,acebrain2026}. 
In recent years, Vision-Language-Action (VLA) models have significantly improved the generalization of embodied agents across diverse tasks by jointly modeling vision, language, and actions~\citep{zitkovich2023rt2,kim2024openvla,black2024pi_0,jiang2023vima}.
Beyond standard VLA modeling, recent studies have further explored world models and agentic frameworks for enhancing long-horizon decision making~\citep{liang2023code,ye2026world}.
For example, CaP-X~\citep{fu2025capx} composes robotic manipulation through code generation, Harness VLA~\citep{zhang2026harness} employs a high-level agent to dynamically orchestrate VLA policies and control primitives, while ASPIRE~\citep{lu2026aspire} constructs and updates robotic skills through interaction. 
Unlike these works that mainly focus on model architectures, policy orchestration, or skill invocation, we focus on the representation and learning of embodied skills themselves, and propose StateSkill, a new skill representation that models embodied interaction experiences as a finite-state machine, enabling skills to dynamically adapt to task progress and environment states.

\subsection{Skill for Large Language Model}
LLM skills abstract historical interaction experiences into reusable behavioral knowledge, allowing agents to transfer previously acquired experience to new tasks~\citep{wu2026co,wang2026skillx,wang2025awm,jiang2026xskill,zhang2026r1}. 
Specifically, Voyager~\citep{wang2024voyager} stores successfully executed programs in a retrievable skill library, ExpeL~\citep{zhao2024expel} summarizes reusable linguistic experience from historical trajectories, while recent works such as SkillRL~\citep{xia2026skillrl} and SkillGen~\citep{ma2026skillgen} further explore automatically constructing and continuously updating skills from successful and failed trajectories. 
Existing LLM skills are typically represented as textual or programmatic task-level knowledge, with limited explicit modeling of continuously changing task states during execution. 
In contrast, StateSkill models a skill as a finite-state machine consisting of task-progress states, state transitions, and state-conditioned guidance, enabling skills to dynamically adapt to changing task and environment states throughout embodied task execution.

\subsection{Reinforcement Learning in LLM Agent}
Reinforcement learning enables LLM agents to directly optimize their decision-making policies through environmental interaction and reward feedback, and has recently been widely adopted for reasoning, tool use, and multi-turn agent tasks~\citep{guo2025deepseek,zhang2025r1,jin2025search,wang2025ragen,feng2026retool,jin2026ecg}.
Methods such as GRPO~\citep{shao2024deepseekmath} optimize model
policies based on group-relative rewards. RL has also been applied
to interactive agent settings through WebRL~\citep{qi2025webrl},
ToolRL~\citep{qian2026toolrl}, and Agent-R1~\citep{cheng2025agent}. Related RL methods mainly optimize model parameters using task outcomes or trajectory-level feedback~\citep{wei2025ddt,kong2026extoken,zhang2025policynewton}, while the internal state structure of long-horizon tasks remains less explicitly exploited. 
Different from existing RL methods for LLM agents, we propose StateGRPO, which introduces task states and state-level feedback into policy optimization, while using newly collected interaction trajectories to further evolve StateSkill, enabling the joint optimization of skill knowledge and model capabilities.

\section{Methodology}

\subsection{StateSkill-Agent Framework}
\label{sec:framework}
As illustrated in Figure~\ref{fig:stateskill-framework},
StateSkill-Agent uses task instructions and current visual
observations to decompose tasks into subtasks and retrieve
corresponding reusable StateSkills. It tracks execution progress
through the explicit states and transitions of each active skill.
Language guidance is introduced progressively as execution advances
through these states, without loading the entire skill description
upfront. After each interaction, a skill-specific gate interprets
observations, executor feedback, and retained facts to update
the active state or return a local skill outcome.

StateSkill-Agent starts with StateSkills constructed by abstracting
shared behavioral patterns from embodied trajectories. Building on these initial skills and the interaction trajectories generated during execution, StateSkill-Evolve refines skill structure and
guidance using new successful and failed state-guided trajectories, while
StateGRPO optimizes StateSkill-Agent's underlying MLLM with
state-conditioned feedback. The updated agent then generates
further interaction experience for skill refinement, forming an
iterative cycle of skill evolution and model optimization.

\subsection{StateSkill: Modeling Embodied Skills as Finite-State Machines}
\label{sec:stateskill}

StateSkill organizes reusable embodied interaction experience into
explicit task-progress states rather than a single task-level
textual description. We formulate a StateSkill as
\begin{equation}
    \mathcal{S}
    =
    \left(
    \mathcal{Q},\mathcal{E},
    \{\kappa_q\}_{q\in\mathcal{Q}}
    \right),
    \qquad
    \kappa_q=(\ell_q,g_q,p_q),
    \label{eq:stateskill-representation}
\end{equation}
where $\mathcal{Q}$ denotes the task-progress states,
$\mathcal{E}$ denotes their transitions, and $\kappa_q$ represents
the state-conditioned knowledge associated with state $q$.
$\ell_q$ provides language guidance for the current
task stage, $g_q$ specifies its transition conditions, and $p_q$
denotes a trajectory-derived action prior. Different skills define their own state vocabularies, evidence requirements, and termination conditions.

\paragraph{StateSkill construction.}
We use GPT-5.5~\citep{openai2026gpt55} to construct reusable
StateSkills from exploration trajectories $\mathcal{D}_{\mathrm{init}}$
through segmentation, cross-task alignment, and knowledge synthesis.
Within a given family of local operations, trajectories are partitioned
into stages according to local objectives and progress conditions.
Compatible stages are aligned into shared states, with task-specific
entities parameterized by functional roles while retaining relevant
relations. Stage ordering provides candidate transitions; recorded
feedback and execution evidence inform their conditions, including
progress, retries, and termination. For each state $q$, we use
supporting segments to synthesize language guidance $\ell_q$, specify
transition conditions $g_q$, and identify action preferences encoded
by $p_q$, yielding $\kappa_q=(\ell_q,g_q,p_q)$ and the initial
StateSkill $\mathcal{S}^{(0)}$.

\paragraph{StateSkill inference.}
Let $q_t$ denote the active state and $h_t$ StateSkill-Agent's
context, including the task instruction, interaction history,
and progressively introduced guidance. The active guidance
$\ell_{q_t}$ conditions the selection of a high-level action category
$c\in\mathcal{A}$. When the MLLM backend exposes action-category
scores, the prior $p_{q_t}$ reweights its decision distribution:
\begin{equation}
    \widetilde{\pi}_{\theta}(c\mid h_t,q_t)
    =
    \frac{
        \pi_{\theta}(c\mid h_t)\,p_{q_t}(c)
    }{
        \sum_{c'\in\mathcal{A}}
        \pi_{\theta}(c'\mid h_t)\,p_{q_t}(c')
    }.
    \label{eq:stateskill-policy}
\end{equation}
The agent generates the corresponding action arguments based on
the selected action category and its current context.
The action vocabulary is fixed within the configured runtime.

After execution, a deterministic skill-specific gate interprets the
returned observations and feedback $e_{t+1}$ together with retained
execution facts $m_t$. For a non-terminal interaction, it updates
\begin{equation}
    (q_{t+1},m_{t+1})
    =
    \delta_{\mathcal S}
    \bigl(q_t,m_t,e_{t+1};g_{q_t}\bigr).
    \label{eq:stateskill-transition}
\end{equation}
The gate can retain the state, follow a valid transition, or return a
local skill outcome to task-level control. Local skill completion does not imply overall task success. On a state transition, the new guide is appended to the context and supersedes earlier guides
as operating guidance; the active action prior changes accordingly.

\subsection{StateSkill-Evolve}
\label{sec:stateskill-evolve}

StateSkill-Evolve uses StateSkill-Agent's experience to refine
how skills represent progress and guide behavior. At iteration
$k$, an offline LLM updater analyzes
$\mathcal{D}^{(k)}=\mathcal{D}_{+}^{(k)}\cup\mathcal{D}_{-}^{(k)}$,
the successful and failed trajectories collected under
$\mathcal{S}^{(k)}$. For each visited state, it compares the active guidance,
transition conditions, and action priors with the recorded
actions and execution evidence. Successful trajectories identify effective behavior
and its conditions of applicability; failed trajectories help identify
missing guidance, inappropriate action preferences, or state distinctions
and recovery paths absent from the current representation. Together, they support revisions grounded in specific states and trajectory segments to preserve effective behavior and address weaknesses.

The updater can revise the state structure $(\mathcal{Q},\mathcal{E})$
and state-conditioned knowledge $\kappa_q=(\ell_q,g_q,p_q)$ to produce
a candidate $\widetilde{\mathcal{S}}^{(k+1)}$. Knowledge updates refine
guidance, transition conditions, and action preferences; structural
updates adjust the state decomposition and transition graph when the
execution evidence warrants them. Each candidate remains a finite-state
skill over the available action and feedback interfaces. Revisions are tied to supporting evidence and an expected behavioral
effect, with unaffected state knowledge retained to preserve useful
experience from earlier iterations.

The candidate and current StateSkill are evaluated on the same
validation task instances with the MLLM checkpoint and execution
configuration held fixed. After consistency and behavioral checks, the candidate is
accepted only if it improves aggregate task success; otherwise,
$\mathcal{S}^{(k)}$ is retained. Validation seeds are separate from the
unseen seeds reserved for final evaluation. This procedure turns
state-level revision hypotheses into execution-tested skill updates.

\subsection{StateGRPO}
\label{sec:stategrpo}

StateSkill-Evolve refines skills, while StateGRPO optimizes
StateSkill-Agent's MLLM using task success and state-wise
execution efficiency. At iteration $k$, we collect a group of $G$
trajectories $\{\tau_i\}_{i=1}^{G}$ for a fixed task and environment
seed through StateSkill-guided execution with the MLLM policy
$\pi_{\theta^{(k)}}$. Training seeds are disjoint from those used
for final evaluation.

Let $s_i\in\{0,1\}$ indicate task success for trajectory $\tau_i$,
and let $T_{i,q}$ denote the total number of interaction turns
spent in state $q$, including repeated visits.
Let $\mathcal{I}_q^{+}$ denote the indices of successful trajectories
in the group that visit state $q$. For
$|\mathcal{I}_q^{+}|\geq 2$, we define the state-specific reference:
\begin{equation}
    \bar{T}_{q}
    =
    \frac{1}{|\mathcal{I}_{q}^{+}|}
    \sum_{j\in\mathcal{I}_{q}^{+}} T_{j,q}.
    \label{eq:state-reference}
\end{equation}
This avoids comparing absolute turn counts across states with
different execution requirements.

Let $\mathcal{C}_i$ contain the states visited by $\tau_i$ for
which $|\mathcal{I}_q^{+}|\geq 2$. For
$\mathcal{C}_i\neq\varnothing$, we define
\begin{equation}
    R_{\mathrm{state}}(\tau_i)
    =
    \frac{s_i}{|\mathcal{C}_i|}
    \sum_{q\in\mathcal{C}_i}
    \frac{\bar{T}_{q}}{T_{i,q}},
    \label{eq:state-reward}
\end{equation}
and set $R_{\mathrm{state}}(\tau_i)=0$ when
$\mathcal{C}_i$ is empty. Failed trajectories receive zero state
reward, while successful ones are rewarded for using fewer turns
relative to the reference in each comparable state.
The comparison does not require identical state sequences
across trajectories.

We combine task success with state efficiency and compute
group-relative advantages:
\begin{equation}
    R_i = s_i + R_{\mathrm{state}}(\tau_i),
    \qquad
    \widehat{A}_i =
    \frac{R_i-\mu_R}{\sigma_R+\epsilon},
    \label{eq:stategrpo-advantage}
\end{equation}
where $\mu_R$ and $\sigma_R$ are the mean and standard deviation
of rewards within the trajectory group, and $\epsilon>0$
ensures numerical stability. Differences in state efficiency can
provide a learning signal even when all trajectories in a group
succeed and therefore receive the same task-success reward.

We optimize the MLLM policy using a clipped GRPO
objective~\citep{shao2024deepseekmath} with advantages
$\widehat{A}_i$. Truncated importance weighting accounts for
differences between the rollout and optimization distributions.
The current StateSkills remain fixed throughout each policy update.

We then collect new StateSkill-guided trajectories with the updated
policy $\pi_{\theta^{(k+1)}}$, providing fresh successful and failed
interaction experience for StateSkill-Evolve.
This alternates model optimization with skill refinement, enabling
reusable skill knowledge and model capabilities to co-evolve.
\section{Experiments}

\subsection{Experimental Setup}
We evaluate StateSkill-Agent on LIBERO-PRO (80 tasks)~\citep{liu2023libero,zhou2025liberopro},
RoboCasa365 (Target50)~\citep{nasiriany2026robocasa365}, and five
real-world manipulation tasks, using GPT-5.5~\citep{openai2026gpt55}
or Qwen3.5-9B~\citep{qwen2026qwen35} as the high-level MLLM.
For LIBERO-PRO, we use 10 environment seeds per task; for
RoboCasa365, we use 10 seeds per atomic task and five per
composite task. Each real-world task is evaluated over 20 trials.
For low-level execution, we use $\pi_{\mathrm{RLinf}}$ for
LIBERO-PRO, RLDX-1 for RoboCasa365, and $\pi_{0.5}$ for
real-world tasks. VLA executors remain frozen; StateGRPO fine-tuning (FT)
updates only Qwen3.5-9B.
We report success rates (\%), with Overall computed as the
unweighted mean of per-task success rates, giving each task
equal weight regardless of its number of evaluation episodes.
Simulation evaluation seeds are held out from skill construction,
skill evolution, candidate validation, and MLLM training.

\begin{table}[!htbp]
\centering
\caption{
\textbf{Results on LIBERO-PRO.}
Success rates (\%) under task (T) and swap (S) perturbations.
Gains (pp) are relative to the corresponding baseline;
``--'' means unavailable.
}
\label{tab:libero_pro}

\resizebox{\textwidth}{!}{%
\begin{tabular}{lccccccccc}
\toprule
\textbf{Method}
& \textbf{Spat-T} & \textbf{Spat-S}
& \textbf{Obj-T} & \textbf{Obj-S}
& \textbf{Goal-T} & \textbf{Goal-S}
& \textbf{L10-T} & \textbf{L10-S}
& \textbf{Overall} \\
\midrule

\multicolumn{10}{c}{\textit{Vision-Language-Action Models}} \\
\midrule

OpenVLA~\citep{kim2024openvla}
& 0.0 & 0.0 & 0.0 & 0.0 & 0.0 & 0.0 & 0.0 & 0.0 & \centeredoverallscore{0.0}{} \\

$\pi_0$~\citep{black2024pi_0}
& 0.0 & 0.0 & 0.0 & 2.0 & 0.0 & 0.0 & 0.0 & 0.0 & \centeredoverallscore{0.3}{} \\

$\pi_{0.5}$~\citep{intelligence2025pi_}
& 1.0 & 20.0 & 1.0 & 17.0 & 2.0 & 38.0 & 1.0 & 8.0 & \centeredoverallscore{11.0}{} \\

MolmoAct~\citep{lee2025molmoact}
& 0.0 & 0.0 & 0.0 & 6.0 & 0.0 & 0.0 & 6.0 & 0.0 & \centeredoverallscore{1.5}{} \\

NORA~\citep{hung2025nora}
& 0.0 & 0.0 & 0.0 & 0.0 & 0.0 & 0.0 & 0.0 & 0.0 & \centeredoverallscore{0.0}{} \\

X-VLA~\citep{zheng2026xvla}
& 0.0 & 0.0 & 8.0 & 2.0 & 9.0 & 1.0 & 10.0 & 0.0 & \centeredoverallscore{3.8}{} \\

AtomVLA~\citep{sun2026atomvla}
& 1.0 & 16.0 & 0.0 & 10.0 & 11.0 & 2.0 & 9.0 & 1.0 & \centeredoverallscore{6.3}{} \\

$\pi_{\mathrm{RLinf}}$~\citep{zang2025rlinf}
& 42.0 & 59.0 & 71.0 & 78.0 & 45.0 & 42.0 & 49.0 & 14.0 & \centeredoverallscore{50.0}{} \\

\midrule
\multicolumn{10}{c}{\textit{Code-as-Policy Agents}} \\
\midrule

CaP-X~\citep{fu2025capx}
& 14.0 & 12.0 & 18.0 & 22.0 & 17.0 & 26.0 & -- & -- & \centeredoverallscore{18.2}{} \\

RATS~\citep{zhang2026playful}
& 31.0 & 29.0 & 63.0 & 61.0 & 36.0 & 43.0 & -- & -- & \centeredoverallscore{43.8}{} \\

RHO~\citep{elmaaroufi2026rho}
& 61.6 & 53.0 & 37.2 & 12.2 & 55.6 & 50.6 & -- & -- & \centeredoverallscore{45.0}{} \\

ASPIRE~\citep{lu2026aspire}
& 60.0 & 51.0 & 95.0 & 98.0 & 45.0 & 81.0 & -- & -- & \centeredoverallscore{71.7}{} \\

\midrule

GPT-5.5 (Baseline)
& 40.0 & 46.0 & 55.0 & 29.0 & 41.0 & 35.0 & 11.0 & 12.0 & \centeredoverallscore{33.6}{} \\

Harness VLA~\cite{zhang2026harness}
& 81.0 & 69.0 & 94.0 & 91.0 & 75.0 & 66.0 & 52.0 & 49.0 & \centeredoverallscore{72.1}{} \\

\rowcolor{blue!4}
StateSkill-Agent
& 83.0 & 72.0 & 95.0 & 92.0 & 77.0 & 68.0 & 54.0 & 51.0
& \centeredoverallscore{74.0}{40.4} \\

\midrule

Qwen3.5-9B (Baseline)
& 24.0 & 19.0 & 17.0 & 22.0 & 11.0 & 14.0 & 7.0 & 8.0 & \centeredoverallscore{15.3}{} \\

Harness VLA~\cite{zhang2026harness}
& 36.0 & 22.0 & 27.0 & 31.0 & 40.0 & 26.0 & 11.0 & 13.0 & \centeredoverallscore{25.8}{} \\

\rowcolor{blue!4}
StateSkill-Agent
& 79.0 & 52.0 & 89.0 & 91.0 & 69.0 & 68.0 & 50.0 & 49.0
& \centeredoverallscore{68.4}{53.1} \\

\rowcolor{blue!4}
StateSkill-Agent w/ FT
& 82.0 & 56.0 & 90.0 & 91.0 & 72.0 & 70.0 & 54.0 & 52.0
& \centeredoverallscore{70.9}{55.6} \\

\bottomrule
\end{tabular}%
}
\end{table}

\begin{table}[!htbp]
\centering
\caption{
\textbf{Results on RoboCasa365 (Target50).}
Success rates (\%) on atomic and composite tasks.
Gains (pp) are relative to the corresponding baseline.
}
\label{tab:RoboCasa365}

\resizebox{\textwidth}{!}{%
\begin{tabular}{lcccc}
\toprule
\textbf{Method}
& \textbf{Atomic-Seen}
& \textbf{Composite-Seen}
& \textbf{Composite-Unseen}
& \textbf{Overall} \\
\midrule

PRTS~\citep{zhang2026prts}
& 66.3 & 30.3 & 18.8 & \centeredoverallscore{39.6}{} \\

RLDX-1~\citep{kim2026rldx1}
& 67.6 & 27.9 & 8.5 & \centeredoverallscore{36.0}{} \\

WorldDreamer~\citep{wang2024worlddreamer}
& 66.3 & 26.7 & 9.0 & \centeredoverallscore{35.3}{} \\

GR00T N1.5~\citep{bjorck2025gr00tn15}
& 50.7 & 14.8 & 2.7 & \centeredoverallscore{23.9}{} \\

$\pi_{0.5}$~\citep{intelligence2025pi_}
& 39.6 & 7.1 & 1.2 & \centeredoverallscore{16.9}{} \\

$\pi_0$~\citep{black2024pi_0}
& 34.6 & 6.1 & 1.1 & \centeredoverallscore{14.8}{} \\

\midrule

GPT-5.5~(Baseline)
& 54.4 & 33.8 & 8.8 & \centeredoverallscore{33.2}{} \\

Harness VLA~\citep{zhang2026harness}
& 91.7 & 56.3 & 13.8 & \centeredoverallscore{55.4}{} \\

\rowcolor{blue!4}
StateSkill-Agent
& 94.4 & 58.8 & 16.3 & \centeredoverallscore{58.0}{24.8} \\

\midrule

Qwen3.5-9B~(Baseline)
& 22.8 & 11.3 & 2.5 & \centeredoverallscore{12.6}{} \\

Harness VLA~\citep{zhang2026harness}
& 52.2 & 16.3 & 3.8 & \centeredoverallscore{25.2}{} \\

\rowcolor{blue!4}
StateSkill-Agent
& 72.8 & 37.5 & 10.0 & \centeredoverallscore{41.4}{28.8} \\

\rowcolor{blue!4}
StateSkill-Agent w/ FT
& 75.6 & 40.0 & 11.3 & \centeredoverallscore{43.6}{31.0} \\

\bottomrule
\end{tabular}%
}
\end{table}

\subsection{Main Results on Simulation Benchmarks}

Tables~\ref{tab:libero_pro} and~\ref{tab:RoboCasa365} summarize the simulation results. With Qwen3.5-9B, StateSkill-Agent w/ FT achieves overall success rates of 70.9\% on LIBERO-PRO and 43.6\% on RoboCasa365, outperforming Harness VLA by 45.1 and 18.4 percentage points, respectively. On LIBERO-PRO, improvements span all four task families under both task and swap perturbations. On RoboCasa365, gains are concentrated in atomic-seen and composite-seen tasks, while composite-unseen success increases from 3.8\% to 11.3\%, indicating that unseen compositions remain challenging. With GPT-5.5, StateSkill-Agent also improves over Harness VLA by 1.9 and 2.6 percentage points on the two benchmarks, respectively, showing that the benefits extend beyond the Qwen3.5-9B configuration.

\subsection{Real-World Experiments}
We investigate sim-to-real skill transfer on five related real-world
manipulation tasks using simulation-derived StateSkills, preserving
their finite-state structure and adapting only environment-dependent
components. We evaluate each embodied MLLM agent with and without
StateSkill on the same set of tasks, using a frozen $\pi_{0.5}$
policy for low-level execution. As shown in Table~\ref{tab:real_world},
StateSkill increases overall success from 60.0\% to 76.0\% for
Qwen3.5-9B and from 75.0\% to 86.0\% for GPT-5.5, yielding gains
of 16.0 and 11.0 percentage points, respectively. The gains span
all five tasks for Qwen3.5-9B and four for GPT-5.5, whose success
rate on Object Sorting remains unchanged at 100.0\%. On Flower
Insertion, Qwen3.5-9B improves from 30.0\% to 50.0\%, while
GPT-5.5 improves from 55.0\% to 70.0\%, yielding gains of 20.0
and 15.0 percentage points, respectively. These results support
sim-to-real StateSkill transfer with environment-specific adaptation.

\begin{table}[!htbp]
\centering
\caption{
\textbf{Real-world manipulation results.}
Success rates (\%) on five real-world tasks.
GPT-5.5 and Qwen3.5-9B are used as baselines.
}
\label{tab:real_world}

\resizebox{\textwidth}{!}{%
\begin{tabular}{lcccccc}
\toprule
\textbf{Method}
& \textbf{Block Stacking}
& \textbf{Flower Insertion}
& \textbf{Object Sorting}
& \textbf{Pen Placement}
& \textbf{Cup Nesting}
& \textbf{Overall} \\
\midrule

GPT-5.5 (Baseline)
& 80.0 & 55.0 & 100.0 & 55.0 & 85.0 & \centeredoverallscore{75.0}{} \\

\rowcolor{blue!4}
GPT-5.5 + StateSkill
& 90.0 & 70.0 & 100.0 & 75.0 & 95.0
& \centeredoverallscore{86.0}{11.0} \\

\midrule

Qwen3.5-9B (Baseline)
& 70.0 & 30.0 & 70.0 & 60.0 & 70.0 & \centeredoverallscore{60.0}{} \\

\rowcolor{blue!4}
Qwen3.5-9B + StateSkill
& 80.0 & 50.0 & 95.0 & 75.0 & 80.0
& \centeredoverallscore{76.0}{16.0} \\

\bottomrule
\end{tabular}%
}
\end{table}

\subsection{Ablation Studies and Discussion}

\begin{wraptable}{r}{0.5\textwidth}
\vspace{-\baselineskip}
\setlength{\abovecaptionskip}{0pt}
\centering
\caption{
\textbf{Ablation studies on LIBERO-PRO.}
Success rates (\%) with Qwen3.5-9B.
Rows progressively add the indicated components.
}
\label{tab:ablation}

\resizebox{\linewidth}{!}{%
\begin{tabular}{cccc}
\toprule
\textbf{StateSkill}
& \textbf{StateSkill-Evolve}
& \textbf{StateGRPO}
& \textbf{Overall} \\
\midrule

-- & -- & -- & 15.3 \\

\checkmark & -- & -- & 63.5 \\

\checkmark & \checkmark & -- & 68.4 \\

\rowcolor{blue!4}
\checkmark & \checkmark & \checkmark & \textbf{70.9} \\

\bottomrule
\end{tabular}%
}
\end{wraptable}

\paragraph{Ablation studies.}
Table~\ref{tab:ablation} examines the progression from
finite-state skill construction to skill evolution and
skill--model co-evolution on LIBERO-PRO with Qwen3.5-9B.
StateSkill achieves 63.5\% success by organizing interaction
experience into task-progress states with state-conditioned
guidance.
Further with StateSkill-Evolve, we build our \textbf{StateSkill-Agent}, which increases success to 68.4\%
(+4.9 percentage points) through retrospective refinement
using new successful and failed state-guided trajectories,
supporting continued skill improvement beyond initial
construction.
Further involving StateGRPO, our final model, \textbf{StateSkill-Agent w/ FT}, raises success to 70.9\%
(+2.5 percentage points) by alternating skill evolution
with state-conditioned GRPO updates to the embodied MLLM agent.

\paragraph{Comparison with LLM skills.}
Table~\ref{tab:llm_skills} compares StateSkill-Agent with
Harness VLA and representative LLM skill methods using the
same Qwen3.5-9B backbone. During task execution, the agent reuses a shared skill library
across tasks without retrieving task-matched reference trajectories.
StateSkill-Agent w/ FT achieves 70.9\% success versus 25.8\%
for Harness VLA, with reported skill length of 11.0K versus
169.4K and token usage of 0.44M versus 3.62M. Compared with
our reproductions of XSkill~\citep{jiang2026xskill},
SkillGen~\citep{ma2026skillgen}, and
Trace2Skill~\citep{ni2026trace2skill}, it achieves higher
success rates with fewer interaction turns and lower token usage.

\begin{figure}[!htbp]
    \centering
    \includegraphics[width=\linewidth]{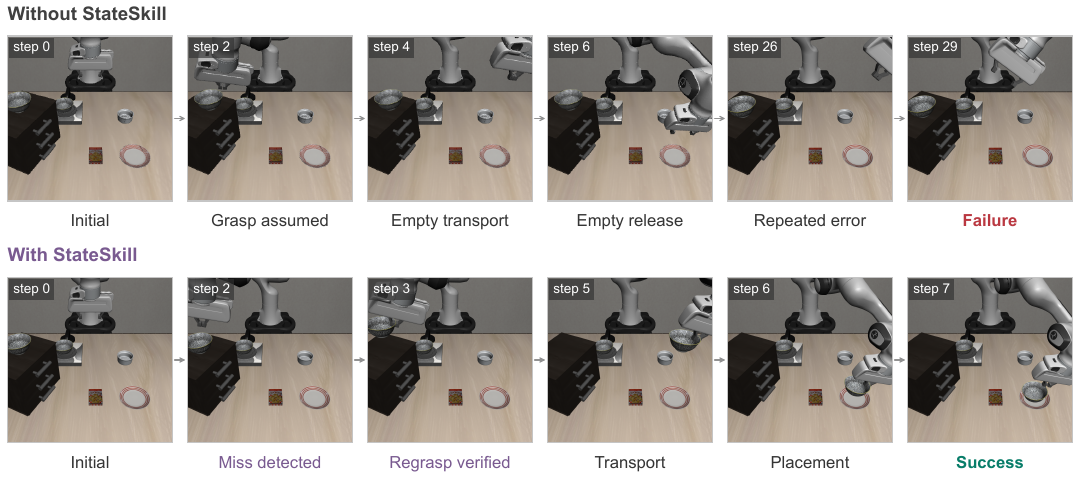}
    \caption{
    \textbf{Qualitative results of StateSkill.}
    Without StateSkill, the agent incorrectly assumes a successful grasp
    and continues with empty transport and release, causing repeated errors
    and eventual failure in a pick-and-place task.
    With StateSkill, the missed grasp is detected, enabling the agent to
    regrasp, verify the updated task state, and successfully complete the task.
    }
    \label{fig:stateskill-ablation}
\end{figure}

\paragraph{Qualitative analysis.}
Figure~\ref{fig:stateskill-ablation} contrasts pick-and-place
execution with and without StateSkill after a missed grasp.
Without StateSkill, the agent assumes grasping succeeded and
proceeds with empty transport and release, causing task failure.
With StateSkill, it detects the missed grasp, retries grasping,
verifies the updated state, and completes the task.
This example illustrates how task-state verification supports
recovery instead of allowing failed actions to be treated as
completed stages.

\vspace{-7pt}

\begin{table}[!htbp]
\centering
\caption{
\textbf{Comparison with LLM skills on LIBERO-PRO.}
All methods use Qwen3.5-9B as the baseline.
Superscripts in the shaded column indicate improvements over Harness VLA:
percentage-point gains in success rate and relative reductions (\%) in other metrics.
}
\label{tab:llm_skills}

\vspace{4pt}
\papertablestyle
\setlength{\tabcolsep}{3pt}
\renewcommand{\arraystretch}{1.12}

\resizebox{\textwidth}{!}{%
\begin{tabular}{@{}lccccc>{\columncolor{blue!4}}c}
\toprule
\textbf{Metric}
& \textbf{Harness VLA}
& \textbf{XSkill}
& \textbf{SkillGen}
& \textbf{Trace2Skill}
& \textbf{StateSkill-Agent}
& \textbf{StateSkill-Agent w/ FT} \\
\midrule

\textbf{Success Rate (\%) $\uparrow$}
& 25.8 & 23.6 & 28.9 & 31.5 & 68.4
& \makebox[2.4em][r]{\textbf{70.9}}%
  \makebox[2.7em][l]{\,$^{\uparrow45.1}$} \\

\textbf{Skill Length (K) $\downarrow$}
& 169.4 & 186.7 & 104.3 & 231.5 & 11.0
& \makebox[2.4em][r]{\textbf{11.0}}%
  \makebox[2.7em][l]{\,$^{\downarrow93.5\%}$} \\

\textbf{Interaction Turns $\downarrow$}
& 74.96 & 76.80 & 69.50 & 66.10 & 23.10
& \makebox[2.4em][r]{\textbf{21.91}}%
  \makebox[2.7em][l]{\,$^{\downarrow70.8\%}$} \\

\textbf{Token Usage (M) $\downarrow$}
& 3.62 & 3.74 & 3.05 & 3.22 & 0.47
& \makebox[2.4em][r]{\textbf{0.44}}%
  \makebox[2.7em][l]{\,$^{\downarrow87.8\%}$} \\

\bottomrule
\end{tabular}%
}
\end{table}

\vspace{-7pt}
\section{Conclusion}
\vspace{-3pt}

This paper presents StateSkill, a new skill representation for embodied agents, which introduces the concept of ``finite-state machines'' into skill modeling for effective and efficient skill construction and learning. The core idea of StateSkill is to model embodied interaction experiences as a finite-state machine, where a skill is organized around a set of task-progress states, state transitions, and state-conditioned guidance. By explicitly conditioning skills on progress states, StateSkill enables an embodied agent to track the continuously changing states of the agent, environment, and task and dynamically adjust its skills in a state-conditioned manner as the task progresses, while keeping skill contexts compact and stage-relevant. Furthermore, we extend StateSkill with an iterative experience-driven evolution mechanism, termed StateSkill-Evolve, which retrospectively analyzes new successful and failed trajectories and refines the state-conditioned guidance, state transition gates, and action priors. In addition, we introduce StateGRPO, which integrates StateSkill evolution and model optimization into a unified iterative learning loop. With StateSkill, StateSkill-Evolve, and StateGRPO, we train our model, StateSkill-Agent. Extensive experiments demonstrate the effectiveness of our proposed methods and trained model across various embodied tasks.

\FloatBarrier

\addtolength{\bibsep}{-0.6pt}
\bibliography{references}
\bibliographystyle{iclr2027_conference}

\end{document}